\documentclass[letterpaper, 10 pt, conference]{ieeeconf}  

\IEEEoverridecommandlockouts                              

\usepackage{graphicx} 
\usepackage{subfig}
\usepackage{amsmath} 
\usepackage{url}
\usepackage{units}

\newcommand{\M}[1]{\mathbf{#1}}
\newcommand{\V}[1]{\mathbf{#1}}
\title{Self-supervised Point Set Local Descriptors\\ for Point Cloud Registration
\thanks{This work  was supported by a German Academic Exchange Service (DAAD) scholarship granted to Yijun Yuan.}
}

\author{Yijun Yuan$^{1}$, Jiawei Hou$^{1}$, Andreas N\"uchter$^{2}$ and S\"oren Schwertfeger$^{1}$
\thanks{$^{1}$The authors are with the School of Information Science and Technology, 
	ShanghaiTech University, China.
	{\tt\small [yuanwj,houjw,soerensch]@shanghaitech.edu.cn}}%
\thanks{$^{2}$The author is with the Department of Informatics VII -- Robotics and Telematics, 
	Julius-Maximilians-University W\"urzburg, Germany.
	{\tt\small [andreas.nuechter]@uni-wuerzburg.de}}
}

\begin{document}
\maketitle
\thispagestyle{empty}
\pagestyle{empty}

\begin{abstract}
In this work, we propose to learn local descriptors for point clouds in a self-supervised manner. In each iteration of the training, the input of the network is merely one unlabeled point cloud. 
On top of our previous work, that directly solves the transformation between two point sets in one step without correspondences, the proposed method is able to train from one point cloud, by supervising its self-rotation, that we randomly generate. The whole training requires no manual annotation. In several experiments we evaluate the performance of our method on various datasets and compare to other state of the art algorithms. The results show, that our self-supervised learned descriptor achieves equivalent or even better performance than the supervised learned model, while being easier to train and not requiring labeled data.

\end{abstract}
\section{Introduction}

Point cloud registration (PCR) is an essential task in various applications, such as 3D reconstruction and simultaneous localization and mapping (SLAM). Usually, the accuracy of the calculated transformation will dominate the performance of higher level tasks. Thus, researchers either make back-end optimization on the high level task, such as SLAM~\cite{durrant2006simultaneous} or work on the PCR side. 
In PCR, rigid-transformation is mostly considered. The Iterative Closest Point (ICP) algorithm, which iteratively solves the point correspondences and transformation, is the most famous algorithm in this field and has been widely used. Using correspondence criterion other than closeness promises improvements in point cloud registration, especially if no good initial guess is available. Descriptors on feature points are used for that. However, the correspondence computing with features requires a good distinctiveness of the descriptor, but the performance of different descriptors usually varies on various point sets. The popular hand-crafted detector ISS~\cite{zhong2009shape} and descriptor FPFH~\cite{rusu2009fast} are widely used.

Aside from the handcrafted descriptors, in recent years, with the fast development of image recognition, deep learning comes into the view. Both point-wise supervised models~\cite{zeng20173dmatch,gojcic2019perfect,deng2018ppfnet} and weakly supervised methods~\cite{yew20183dfeat} are proposed to improve the matching performance. 
However, those supervising requires large amounts of labor to label the data. Those algorithms are either getting the correspondence from the matched point clouds~\cite{zeng20173dmatch,gojcic2019perfect,deng2018ppfnet}, which is costly, or they are labeling the inter-point cloud relation~\cite{yew20183dfeat}, which is is inefficient to train. In addition, the existing supervised models usually come with a triplet siamese structure or other loss functions, that are not directly related to the registration. 

Therefore, we wonder if there is some method that does not require any labeling for training. Which means, that the supervised information comes from the same point cloud itself. Thus we call our method, self-supervised learning of point set local descriptors for point cloud registration.

\begin{figure*}[t]
	\centering
	\includegraphics[width=1.\linewidth]{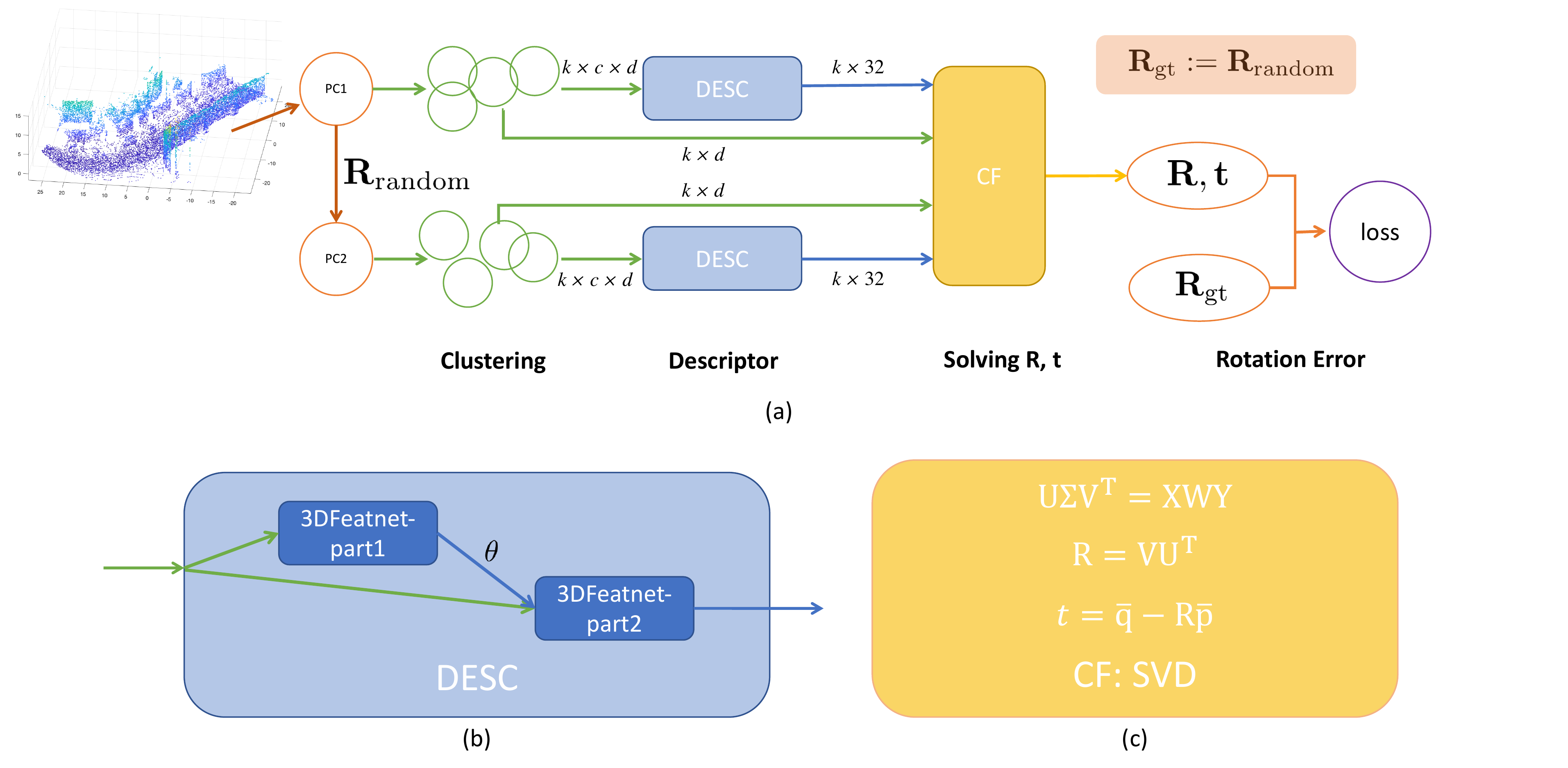}
	\caption{(a) Pipeline of training: Single input point cloud; branching with random rotation; clustering; descriptor; CF algorithm to estimate R, t and rotation error as loss function. (b) Detail of of the descriptor. (c) SVD part of CF.}
	\label{fig:pipe}
\end{figure*}

Self-supervised learning is proposed for utilizing unlabeled data with the success of supervised learning. Producing a dataset with good labels is expensive, while unlabeled data is being generated all the time. The motivation of self-supervised learning is to make use of the large amount of unlabeled data. The main idea of self-supervised learning is to generate the labels from unlabeled data, according to the structure or characteristics of the data itself, and then train on this unsupervised data in a supervised manner. Self-supervised learning is wildly used in representation learning to make a model learn the latent features of the data. This technique is often employed in computer vision~\cite{dosovitskiy2015discriminative, gidaris2018unsupervised, doersch2015unsupervised, zhang2016colorful, donahue2016adversarial}, video processing~\cite{wang2015unsupervised, vondrick2018tracking} and robot control~\cite{jang2018grasp2vec, zhi2019learning, nair2019contextual}. Similarly, we also find a work on self-supervised learning of 3D local features (MortonNet)~\cite{thabet2019mortonnet}. Given a sequence of $n$ points in Morton order, the MortonNet learns to predict the last point from the first $n-1$ points. However, the paper only discusses the application on segmentation.

In this paper, we propose a self-supervised learning model to learn the point cloud local descriptor for registration. The input of the network is merely one raw point cloud for each iteration of training. In our previous work we proposed the Full Connection Form Solution (CF) model~\cite{yuan2020non} to solve the PCR problem non-iteratively in one-step without correspondences. There, handcrafted descriptors, such as FPFH~\cite{rusu2009fast}, are used, which work well when point clouds are sampled from the same distribution. 

The fully connected graph in CF considers the true corresponding point pairs especially important for the registration. So each descriptor should be similar to its possible corresponded points while different to others. Since the feeds are from one single point cloud, our self-supervise setting fits well to the CF module. 

The idea of the network is that we, given a random rotation, feed the original point cloud PC1 and its rotated point cloud PC2 into the neural net to learn the feature, then solve the transformation with CF. Since the whole model is differentiable, with a loss function on the predicted transformation, we propagate the gradient back to the descriptor network. The whole pipeline diagram is shown in Fig.~\ref{fig:pipe}.

We build on the 3DFeatNet~\cite{yew20183dfeat} as the Deep Neural Network body, as it solely feeds in points and does not require additional processing on the data representation.

Experiments on various datasets, i.e.,~\cite{maddern20171, geiger2012we}, demonstrate the performance of our descriptor.

To summarize the major contributions of this paper are:
\begin{itemize}
	\item A new self-supervised learning method for descriptors in point clouds, that requires no manual annotation.
	\item Experiments that show that the self-supervised learned local descriptor has an equivalent performance as the supervised 3DFeatNet.
\end{itemize}

\section{Related work}

This section reviews the technique advances in 3D local point cloud descriptors for registration. It describes handcrafted descriptors and learned models. The handcrafted features usually resort on the geometric correlation inside a local cluster. Borrowing the strong representation ability of Deep Neural Networks, the learned models regress a descriptor function from data.

\subsection{Handcrafted 3D Descriptor}
Point Feature Histogram (PFH) is known as the most typical 3D local descriptor~\cite{rusu2009fast}. It encodes the neighborhood geometrical properties with a multi-dimensional histogram~\cite{rusu2008learning}. For real time application, Fast Point Feature Histogram (FPFH) breaks the full interconnection of neighbors in PFH. Thus it achieves a linear time complexity and gradually becomes the most commonly used handcrafted 3D descriptor~\cite{rusu2009fast}. 

Apart from the descriptors from geometry, spin images (SI)~\cite{johnson1999using} and unique shape context (USC)~\cite{tombari2010unique} split the spatial space into bins and count the number of points in each as a histogram.

\subsection{Learned 3D Local Descriptor}

Focusing on the representation of point clouds fed into the deep neural net (DNN) model, we simply divide the learned descriptors into two classes: the voxel based and the point-set based methods. 

Voxel based methods represent the local region around a keypoint as a volume. The most representative work is 3DMatch~\cite{zeng20173dmatch}. They use a 3D convolutional network to learn the weight from matching and non-matching. With the voxelized smoothed density value (SDV) representation, 3DSmoothNet~\cite{gojcic2019perfect} performs additional pre-processing, local reference frame (LRF), before DNN, to achieve rotation invariance. 

The point set based methods start with the success of PointNet~\cite{qi2017pointnet}, which made it convenient and efficient to learn with DNN. PPFNet~\cite{deng2018ppfnet} and PPF-FoldNet~\cite{deng2018ppf} feed an encoding of points into PointNet to learn the features. However, they are not straight forward, because the encoding requires additional processes such as normal and point pair features. Then 3DFeatNet~\cite{yew20183dfeat}, by feeding in pure points set, models a mapping function from points to descriptors directly. 

3DFeatNet uses whole point clouds instead of local patches as input, which is already different from most works. Thus it merely needs to annotate the point cloud relation instead of point cluster relation. This step saves lots of labor in labeling and tuning on triplet design. Such an advantage resorts to its clustering step which is following PointNet++~\cite{qi2017pointnet++} to separate point clouds into clusters. 
Then to learn the rotation invariance, it additionally uses a network sub-module to learn the orientations, to rotate clusters afterward. During training, an attention value is also extracted for loss function design. 3DFeatNet also has an extra usage. During training, attention is utilized to weight the pairwise distance between clusters of point clouds. However during inference, attention makes no use of the descriptor. So 3DFeatNet creatively uses the attention value to filter out non-interest points from the randomly sampled points, as the 3DFeatNet keypoints. 


\section{Methodology}

\subsection{Problem Statement}
Given two point clouds $\mathbf P$ and $\mathbf Q$, the goal of rigid registration is to solve
\begin{equation}
\min_{\M{R},\M{t}} \sum_{(i,j)\in \mathcal{C}} || \M R \V p_i + \V t - \V q_j ||^2
\end{equation}

where $\V p_i\in \mathbf P | _{i\in{1,\ldots,N}}$, $\V q_j\in\mathbf Q|_{j\in{1,\ldots,M}} $ and $\mathcal{C}$ is the set of correspondences. 

A descriptor for each keypoint $p_i$ is a vector utilized to compute the correspondences with points in the other point cloud. We denote the mapping  from point location to descriptor vector as $f_{\mathcal X}(\V x)$ with point $\V x\in \mathcal X$. So the goal for this paper is to regress such a mapping $f$.

\subsection{Network Architecture}

We demonstrate the pipeline of the training process in Fig.~\ref{fig:pipe}. The DESC module in between is the $f$ we want to extract.

The whole training process consists of four parts. The first two modules, clustering and descriptor, follow 3DFeatNet~\cite{yew20183dfeat}. Next, the CF module~\cite{yuan2020non} solves the transformation of sampled points with descriptors from the two point clouds and the Rotation Matrix Distance Module computes the error between the solved $R$ and $R_{\text{gt}}$, which is then the loss.

\subsubsection{Network Body}
For each of the point clouds, we sample the points from the cloud and then group points surrounding the sampled points into clusters. Following the example of~\cite{yew20183dfeat}, we are using grouping and sampling layers from PointNet++~\cite{qi2017pointnet++} in this paper. With $k$ sampled points as the centers, $k$ clusters are extracted. For that, first all points within radius $r_{\text{cluster}}=2m$ around the sampled point are selected. Then $c$ points are randomly sampled from that set. In the cluster, each point is of dimension $d$, that can be $3$ (xyz), $6$ (xyzrgb) or others. In this paper we use $d=3$, so we are only using the xyz location of the point.

Then the points of those clusters are passed into the descriptor network, which processes each cluster with size $c\times 3$ and outputs their descriptor. The descriptor body consists of two parts, 3DFeatnet-part1 and 3DFeatnet-part2. The 3DFeatnet-part1 learns to generate the orientation to rotate the cluster input of 3DFeatnet-part2. 

3DFeatNet only predicts a 1D rotation to avoid unnecessary equivariances. Thus they only rotate around the gravity axis.
Different from 3DFeatNet~\cite{yew20183dfeat}, that uses a Siamese Network with three branches for triplet input (anchor, positive, negative), we only use two branches with feeding PC1, PC2, as shown in Fig.~\ref{fig:pipe}. The two branches of network share weights.

\subsubsection{Solving $\V R$, $\V t$}

The CF module, as discussed in~\cite{yuan2020non}, solves the registration for point sets with the same distribution in one step. Since PC2 is with a rotation $R_{\text{random}}$ from PC1 that fits for such a requirement, the transformation is solved.

Please note that the location input of CF are the sub-sampled PC1 and PC2 from clustering in Fig. \ref{fig:pipe}.

Given points $\V p \in g( \mathbf P)$, $\V q \in g(\mathbf Q)$ and their descriptors $f_{\mathbf P}(\V p)$ and $f_{\mathbf Q}(\V q)$, the optimization program in the CF module is 
\begin{equation}
\min_{\M R, \V t} \sum_{i=1}^{k_{ \mathbf P}}\sum_{j=1}^{k_{\mathbf Q}} w_{i,j} || \M R \V p_i + \V t - \V q_j|| ^2
\label{eq::cf}
\end{equation}
where 
\begin{equation}
w_{i,j}=e^{-\frac{1}{\alpha} || f_{\mathbf P}(\V p_i) - f_{\mathbf Q}(\V q_j)      ||^2} .
\end{equation}

Let $\mathcal X=\{\V p'_1, \ldots, \V p'_{k_{ \mathbf P} k_{ \mathbf Q}} \}$, $\mathcal Y=\{\V q'_1, \ldots, \V q'_{k_{ \mathbf P}k_{ \mathbf Q}} \}$. $\V p'$, $\V q'$ where the same index indicates the point of $\V p$, $\V q$ in one term of the summation in Eq.~\eqref{eq::cf}. $w$ is correspondingly set.

Then above problem Eq.~\eqref{eq::cf} is transformed into 
\begin{equation}
\min_{\M R, \V t} \sum_{i=1}^{k_{ \mathbf P}k_{ \mathbf Q}}w_{i} || (\M R \V p'_i + \V t) - \V q'_i || ^2
\label{eq::cf_full}
\end{equation}

The transformation $\M R$, $\V t$ in Eq.~\eqref{eq::cf} has a closed form solution using the SVD~\cite{arun1987least}, cf. Fig.~\ref{fig:pipe} (c).


\subsubsection{Loss Function}
The loss function is to supervise the one-step solved rotation~\cite{yuan2020non}.
Given the ground truth transformation $\M R_\text{gt}$, $\V t_\text{gt}$, the loss function is the deviation from the identity matrix~\cite{larochelle2006distance} as follows

\begin{equation}
loss=  ||\M I - \M R\M R_\text{gt}^T||_F .
\label{eq:loss}
\end{equation}

\subsubsection{Self-supervised Learning}
With the above four parts of network components, it merely requires to feed in one raw point cloud to learn for each iteration. 
Given a random rotation, we want to minimize its distance from the solved rotation. 

Since the whole pipeline is differentiable, the parameters in the descriptor network is updated with gradient back-propagation.

We call our model self-supervised learning model, because we generate labels ($\M R_{\text{random}}$) from nothing, and train the unlabeled data in a supervised way. The model is learned from a raw point cloud itself.

\section{Experiments and Results}

Our implementation and experiments are based on top of the open source release\footnote{\url{https://github.com/yewzijian/3DFeatNet}} of 3DFeatNet~\cite{yew20183dfeat}. The Oxford RobotCar dataset~\cite{maddern20171} is used for network training and testing, following the settings of~\cite{yew20183dfeat}. Additionally, the KITTI dataset is also used for testing the model.

\subsection{Datasets}

%
%
\begin{figure}
	\centering
	\subfloat[Oxford Data Used]{
		\label{fig:data:oxford}
		\includegraphics[width=.45\linewidth]{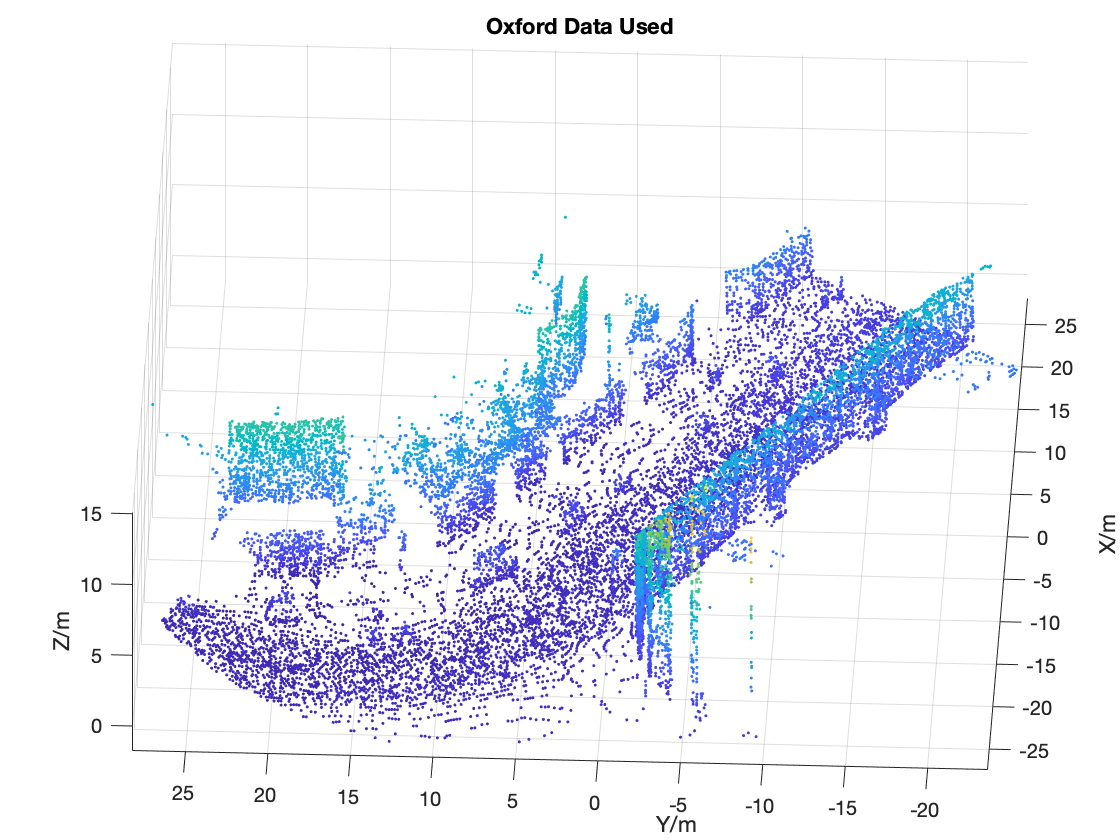}}
	\subfloat[KITTI Data Used]{
		\includegraphics[width=.45\linewidth]{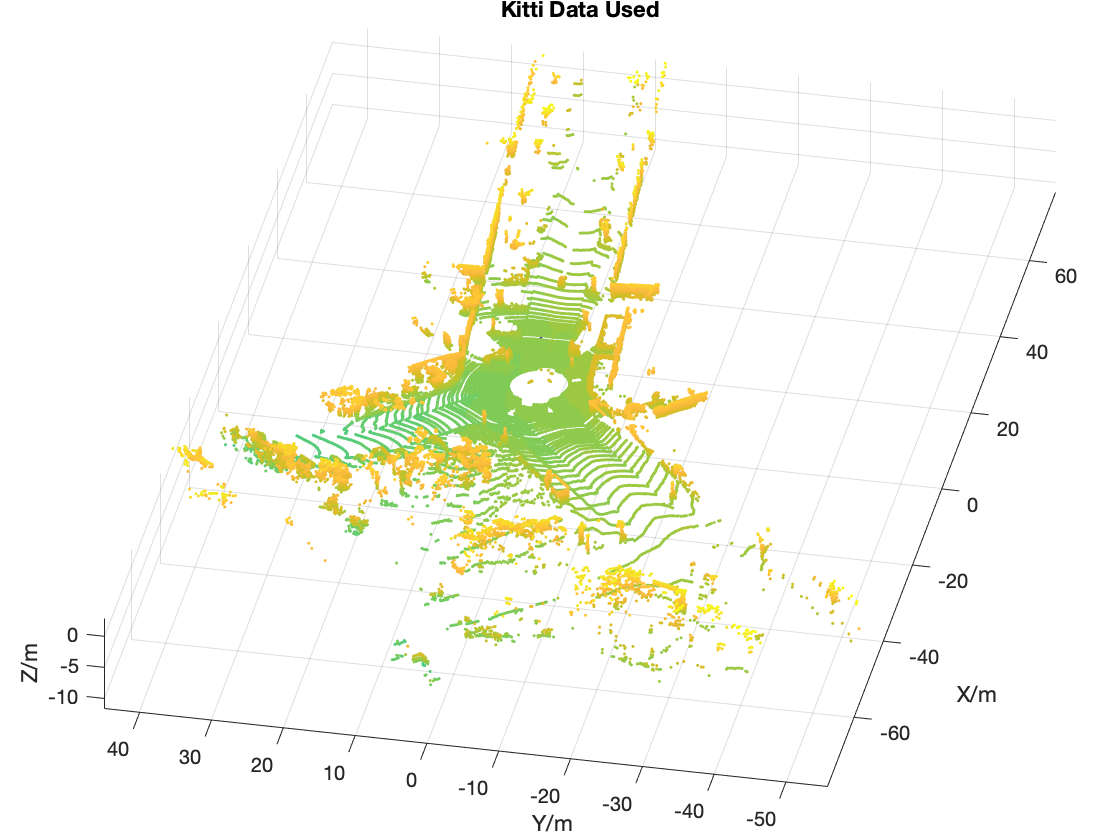}}
	\caption{The processed data utilized in this paper. In Oxford data, the span on X,Y axis is around $50m$. For Kitti data, span on one axis may exceed $100m$.}
	\label{fig:data}
\end{figure}

\subsubsection{Oxford RobotCar Dataset}
The Oxford dataset contains repeat traverses through the Oxford city center from May 2014 to December 2015, that were collected with the Oxford RobotCar platform. 
We use the pre-processed data from 3DFeatNet~\cite{yew20183dfeat}, that has 35 trajectories for training and another 5 trajectories for testing. The points scanned from 2D LIDAR are accumulated into 3D point clouds, according to the GPS/INS poses. Those poses were refined with ICP. 3DFeatNet then downsampled the training point clouds to about $50,000\pm20,000$ points and the test point clouds to exactly 16,384 points. During training, the training point clouds are further randomly subsampled to 4096 points. This way we obtain 21,875 training and 828 testing point cloud sets. 



\subsubsection{KITTI Dataset}
We test our model on the 11 training sequences from KITTI dataset~\cite{geiger2012we} as~\cite{yew20183dfeat}. The parts of KITTI dataset used in the experiments include Velodyne laser point clouds, GPS/INS as ground truth poses and the calibration files. The point clouds are also downsampled with a grid size of \unit[0.2]{m} and obtain 2,369 point clouds in the end.

\begin{table*}[!]
	\centering
	\caption{Registration Error on the Oxford Dataset. The first 8 rows are collected from~\cite{yew20183dfeat}. We obtained the last 5 rows using ISS keypoints, FPFH, 3DFeatNet and our approach. The parameters are the same as in~\cite{yew20183dfeat}.}
	\begin{tabular}{|l| |l|l|l|l|}
		\hline
		& RTE & RRE & Success Rate & Avg \#Iter \\ \hline \hline
		ISS+FPFH~\cite{rusu2009fast} & 0.396 & 1.60 & 92.32\% & 7171 \\ 
		ISS+SI &0.415&1.61&87.45\%& 9888 \\ 
		ISS+USC& 0.324 & 1.22 & 94.02\% & 7084 \\ 
		ISS+CGF& 0.431&1.62&87.36\%&9628 \\
		ISS+3DMatch &0.494&1.78&69.06\%& 9131 \\ 
		ISS+PN++ &0.511&1.88&48.86\%&9904 \\ 
		ISS+3DFeatNet Desc &0.314&1.08&97.66\%&7127 \\ 
		3DFeatNet Kpt+3DFeatNet Desc &0.300&1.07&98.10\%&2940 \\ 
		\hline \hline
		ISS+FPFH~\cite{rusu2009fast} &0.354&1.43&93.37\%&3239 \\ 
		ISS+3DFeatNet Desc~\cite{yew20183dfeat}&0.314 &1.08& 97.66\% & 7126 \\ 
		ISS+Our Desc & 0.311&\textbf{1.01} & \textbf{98.10\%}&5648\\
		3DFeatNet Kpt+3DFeatNet Desc &\textbf{0.304}&1.08&97.66\%&\textbf{3294}\\
		3DFeatNet Kpt+Our Desc &0.310&1.08&97.05\%&3650 \\ \hline
	\end{tabular}
	\label{tab::oxford_registration}
\end{table*}

\begin{table*}[!]
	\centering
	\caption{Registration Error on the KITTI Dataset. The first 6 rows are collected from~\cite{yew20183dfeat}. We obtained the last 4 rows using ISS keypoints, FPFH and our approach.}
	\begin{tabular}{|l| |l|l|l|l|}
		\hline
		& RTE & RRE & Success Rate & Avg \#Iter \\ \hline \hline
		ISS+FPFH~\cite{rusu2009fast} & 0.325 & 1.08 & 58.59\% & 7462 \\ 
		ISS+SI &0.358&1.17&55.92\%& 9219 \\ 
		ISS+USC& 0.262 & 0.83 & 78.24\% & 7873 \\ 
		ISS+CGF&0.233&0.69&87.81\%&7442 \\
		ISS+3DMatch &0.283&0.79&89.12\%&7292 \\ 
		3DFeatNet Kpt+3DFeatNet Desc &0.258&0.57&95.97\%&3798 \\ 
		\hline \hline
		ISS+3DFeatNet Desc& 0.246&0.627& 93.50\%&8311\\ 
		ISS+Our Desc &\textbf{0.215} &\textbf{0.510}&93.50\%&5960 \\
		3DFeatNet Kpt+3DFeatNet Desc &0.264&0.599&\textbf{95.58}\%&4394\\
		3DFeatNet Kpt+Our Desc &0.258&0.570&95.44\%& \textbf{3732} \\ \hline
	\end{tabular}
	\label{tab::kitti_registration}
\end{table*}

\subsection{Setting}
On training, 3DFeatNet feeds the network a triplet of 3 point clouds (anchor, positive, negative) from the Oxford Dataset. For our training of the model, we simply use the one point cloud (anchor).
The clustering and network settings are following 3DFeatNet. The radius of the cluster is $r_{\text{cluster}}=2m$.

Since our work only consists of the descriptor, we use the keypoints from other detectors. The detectors used in this evaluation are ISS~\cite{zhong2009shape} and 3DFeatNet keypoints (kpt). The experiment consists of descriptor matching joint performance with keypoint and geometric registration.
The other baseline descriptors are FPFH~\cite{rusu2009fast}, SI~\cite{johnson1999using}, USC~\cite{tombari2010unique}, CGF~\cite{khoury2017learning}, 3DMatch~\cite{zeng20173dmatch} and 3DFeatNet~\cite{yew20183dfeat}.
In the CF module, we set $\alpha=1$. 

The 3DFeatNet takes 2 epochs  to pretrain 3DFeatNet-part2 and trains whole model 70 epochs with $lr=1e-5$. We use the open released sample Tensorflow~\cite{abadi2016tensorflow} checkpoint to achieve the network weight of 3DFeatNet~\cite{yew20183dfeat}.

During the training of our model, we follow the 3DFeatNet to set the network hyperparameters, such as batch size 6, Adam optimizer and 32 dimension descriptor.
3DFeatNet claims that it is hard to train, while our network is easy to train: Without any pre-training, our model is randomly initialized and saved at iteration 72,500 (same as 3DFeatNet's 20 epochs training) with learning rate $lr=1e-3$.
The provided $R_{\text{random}}$ is generated to only have the Z-axis rotate $\phi\sim \mathcal{N}(0,\sigma_{r}^2)$. In the experiment, we set $\sigma_r=0.6$. In addition, we apply a 3D jitter with $\V \Delta p \sim \mathcal{N}(\V 0, \sigma_p\V I )$ ($\sigma_p=0.01$) for each point in PC1 and PC2. 

On inferencing, the setting of the 3DFeatNet detector such as, $\beta_{attention}$ and $r_{nms}$, follows~\cite{yew20183dfeat}.

\subsection{Evaluations}

To evaluate the performance of our descriptor, we use ISS and 3DFeatNet detector to provide the keypoint, and demonstrate the performance on 
precision and geometric registration.

\begin{figure}
	\centering
	\subfloat[ISS Detector]{
		\includegraphics[width=.5\linewidth]{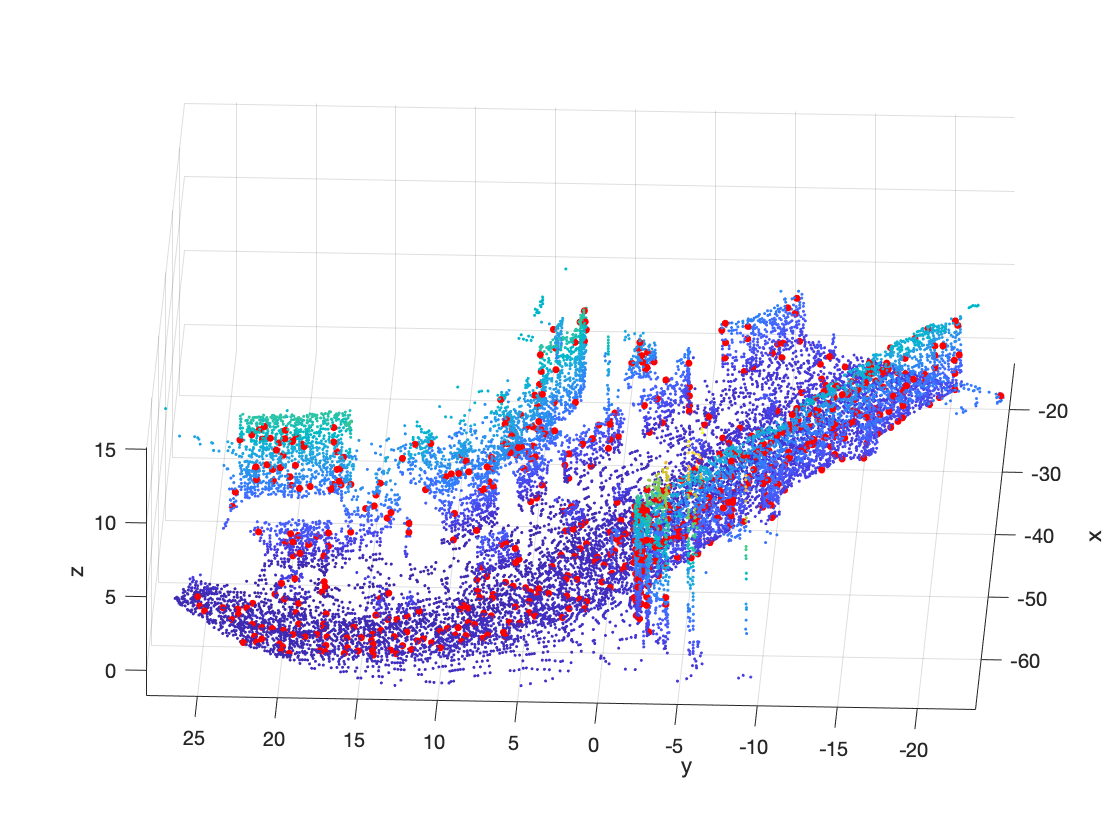}}
	\subfloat[3DFeatNet Detector]{
		\includegraphics[width=.5\linewidth]{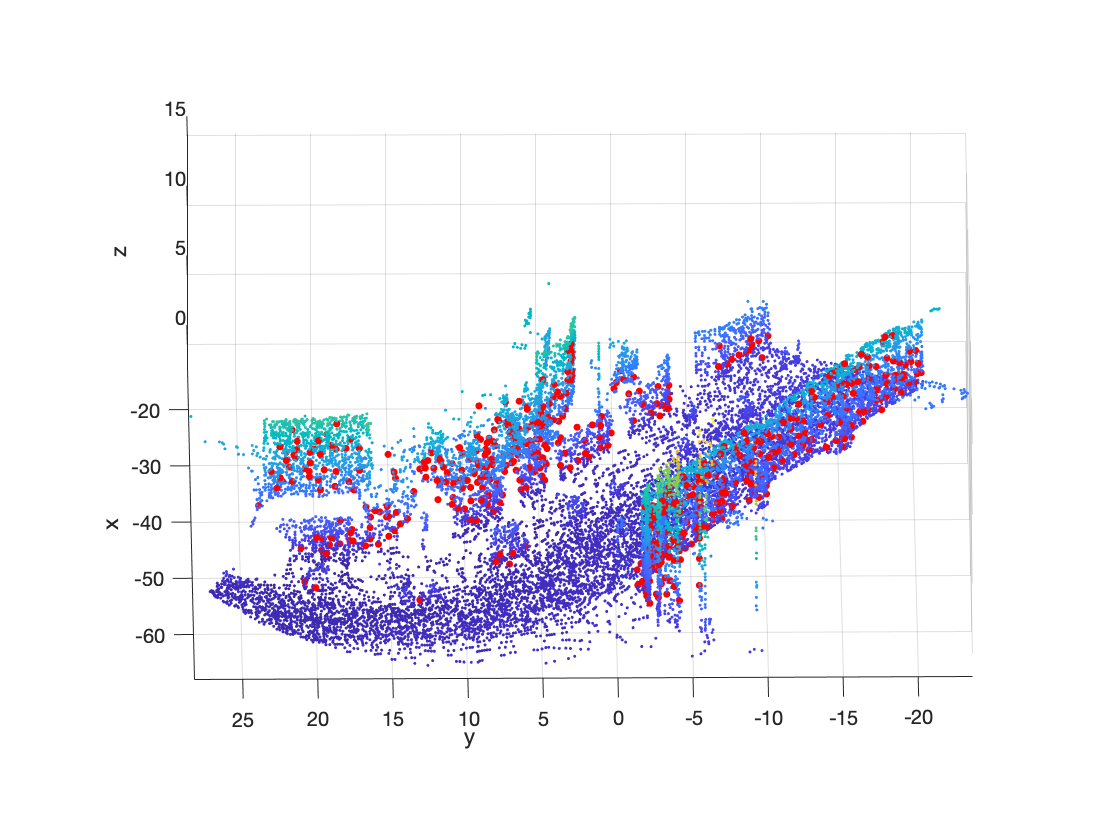}}
	\caption{Keypoint Demo on the same point cloud of Fig. \ref{fig:data:oxford}. Key-points are plotted with red dots on point cloud.}
	\label{fig:keypoint}
\end{figure}

An example of keypoint detection of ISS and 3DFeatNet detector is shown in Fig.~\ref{fig:keypoint}. We observe that the interest points of the ISS are distributed in the whole point cloud while most keypoints of 3DFeatNet detector are on the wall.

\begin{figure}[]
	\centering
	\includegraphics[width=.7\linewidth]{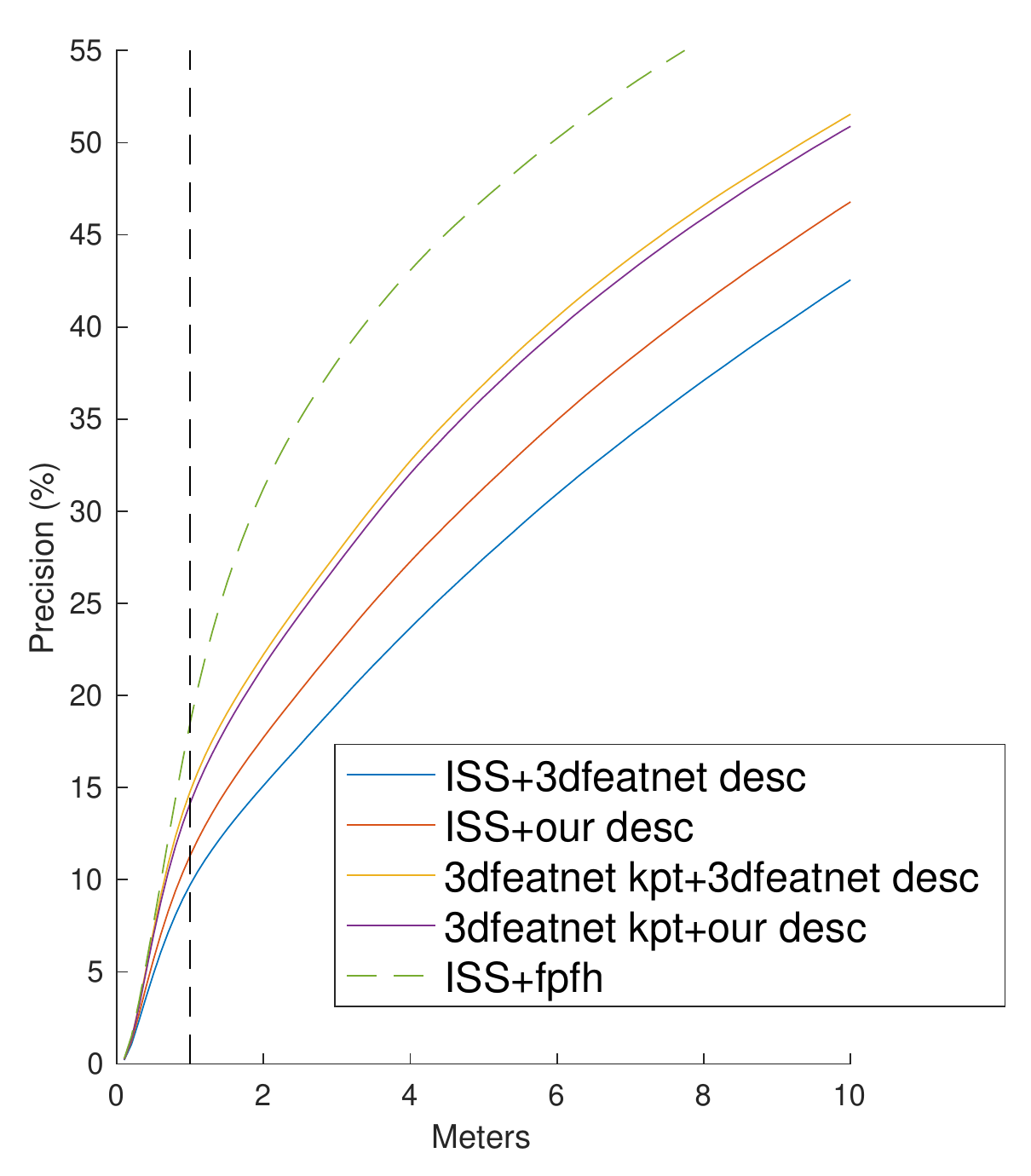}
	\caption{Precision plot for distance between nearest neighbor point and the ground truth location.}
	\label{fig:kp_desc}
\end{figure}

\subsubsection{Precision Test}
Using exhaustive search, this test searches for the nearest descriptor neighbor in the paired model for each keypoint. Then the Euclidean distance between the neighbor and ground truth location is computed. 
We show the plot in Fig.~\ref{fig:kp_desc}. The x-axis is a threshold to consider a pair as correct and the y-axis is the correct proportion.
Comparing to the plot in~\cite{yew20183dfeat}, the tested ISS+FPFH achieved a better result than the plot in~\cite{yew20183dfeat}.
 
For both 3DfeatNet Descriptor and our descriptor, the test with 3DFeatNet kpt works better than ISS kpt. Our proposed  unsupervised model achieved a similar result to 3DFeatNet Desc with 3DfeatNet kpt and a better result on ISS kpt, comparing with 3DFeatNet Desc. We use the $x=1m$ line as a cut. Both 3DFeatNet Desc and our model achieves around $15\%$ precision, which is close to the best score in the record of~\cite{yew20183dfeat}.
\begin{figure*}
	\centering
	\subfloat[Scene 1. ISS+our desc]{
		\includegraphics[width=.25\linewidth]{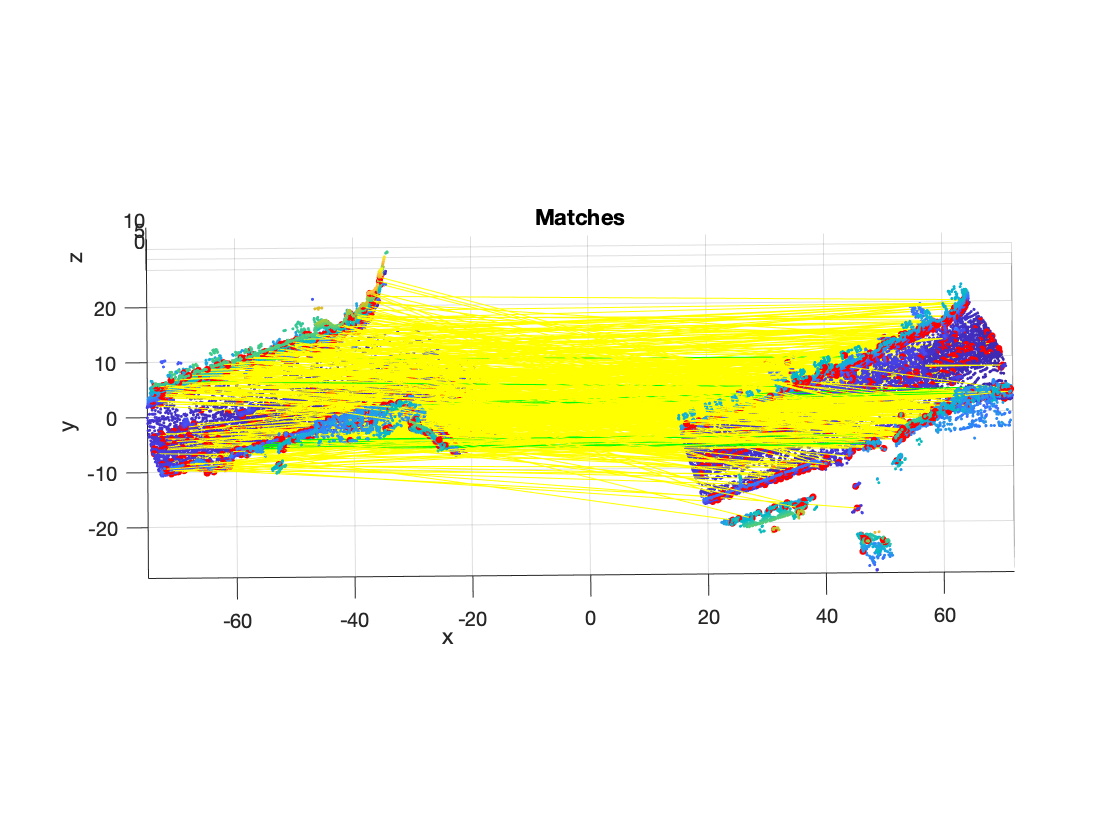}}
	\subfloat[Scene 1. 3DFeatNet kpt+our desc]{
		\includegraphics[width=.25\linewidth]{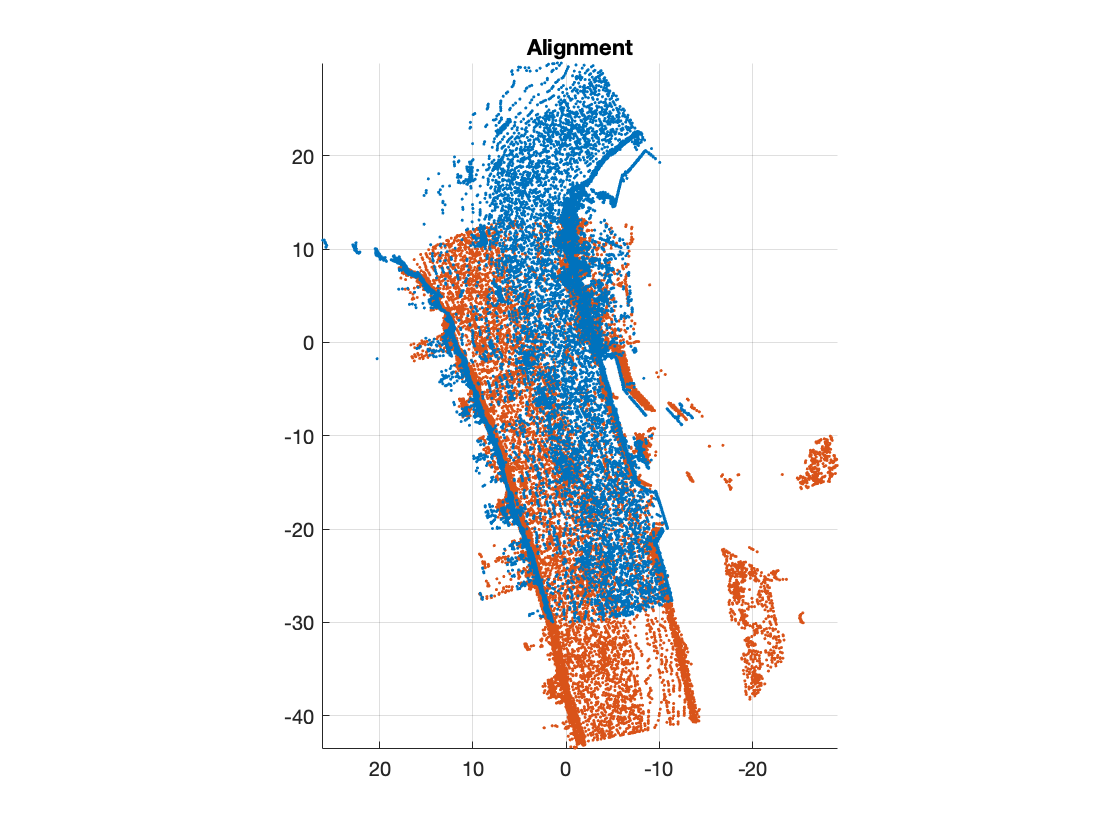}}	
	\subfloat[Scene 2. ISS+our desc]{
		\includegraphics[width=.25\linewidth]{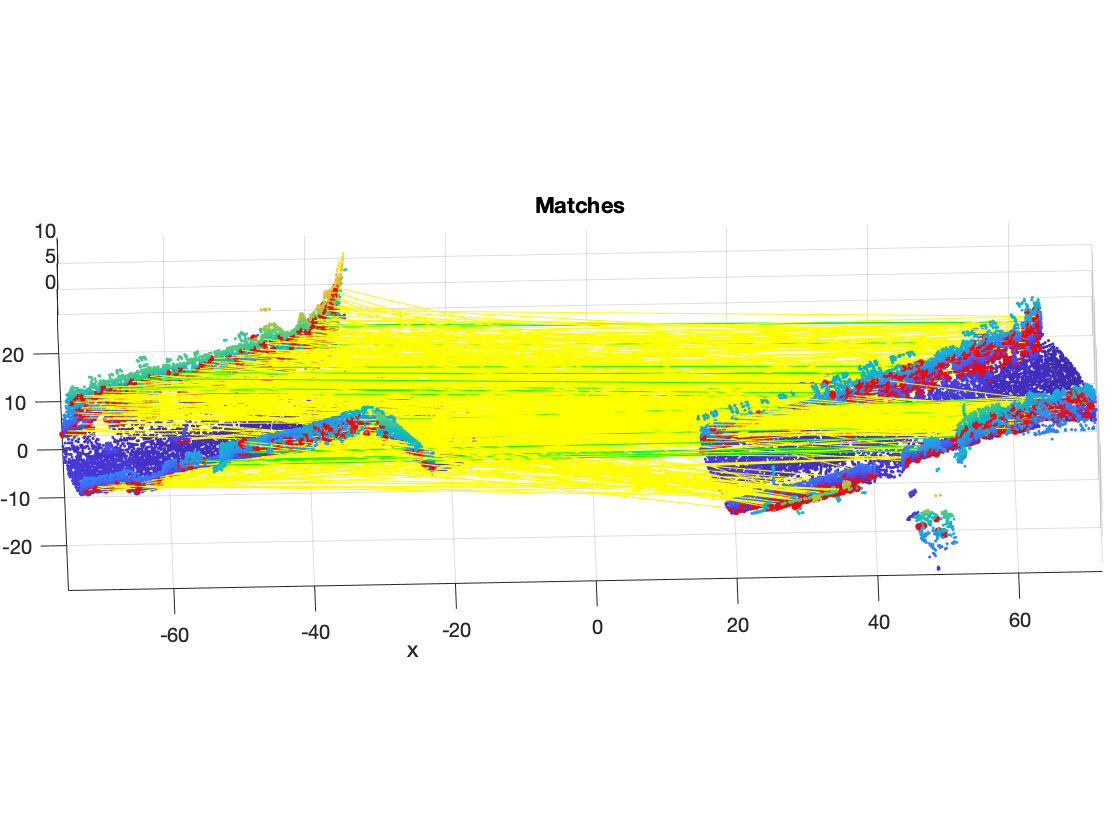}}
	\subfloat[Scene 2. 3DFeatNet kpt+our desc]{
		\includegraphics[width=.25\linewidth]{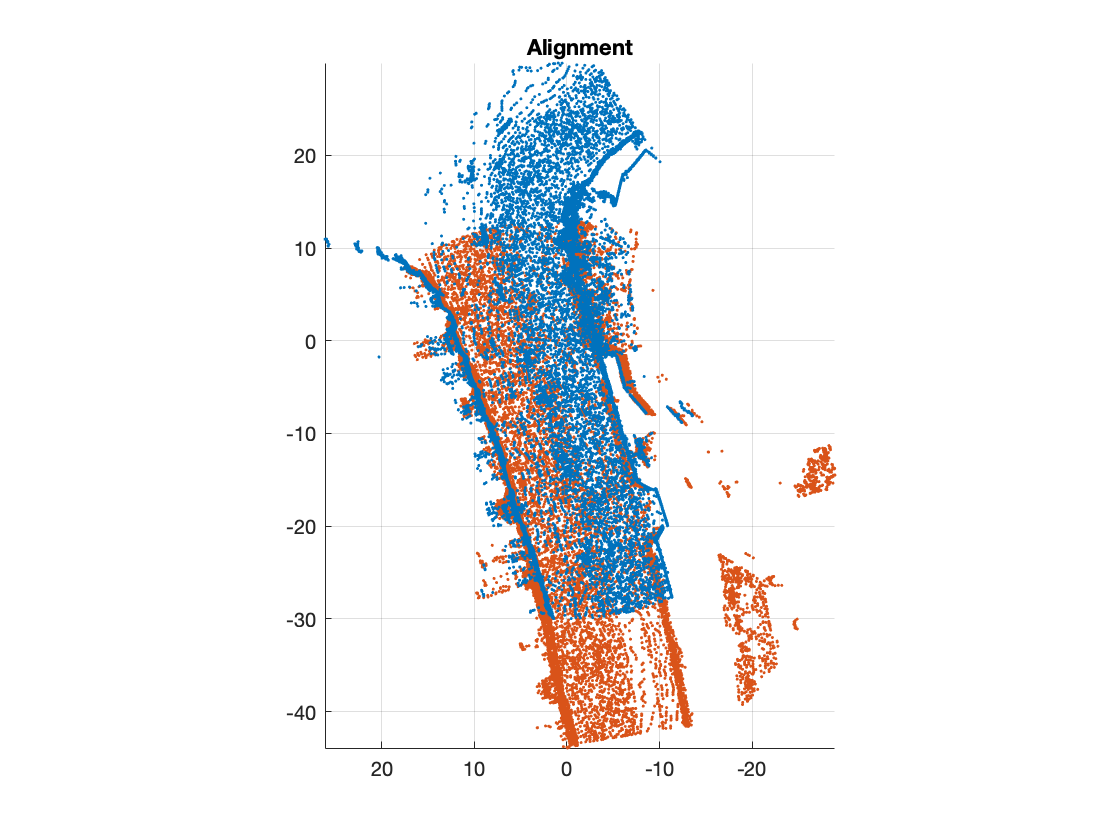}}
	\caption{Oxford data geometric registration. Top view. Yellow dot line shows the matched pairs.}
	\label{fig:match_oxford}
\end{figure*}
\begin{figure*}
	\centering
	\subfloat[Scene 1. ISS+our desc]{
		\includegraphics[width=.25\linewidth]{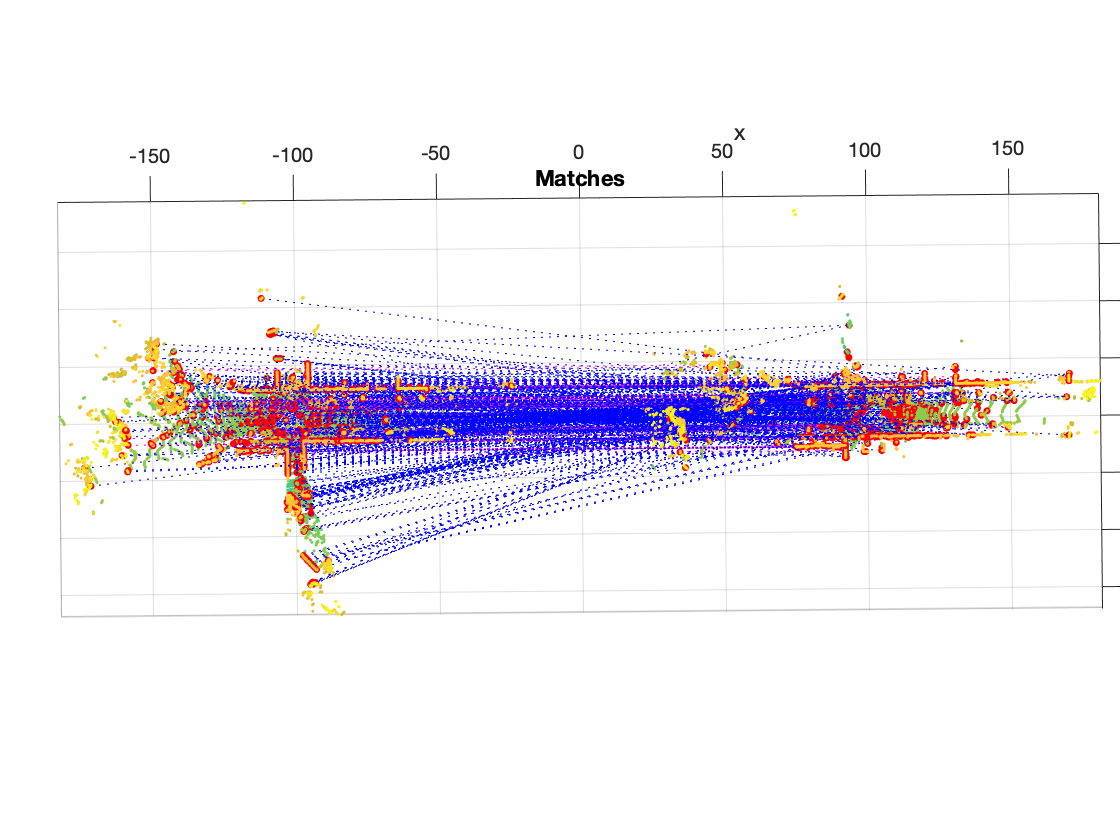}}
	\subfloat[Scene 1. 3DFeatNet kpt+our desc]{
		\includegraphics[width=.25\linewidth]{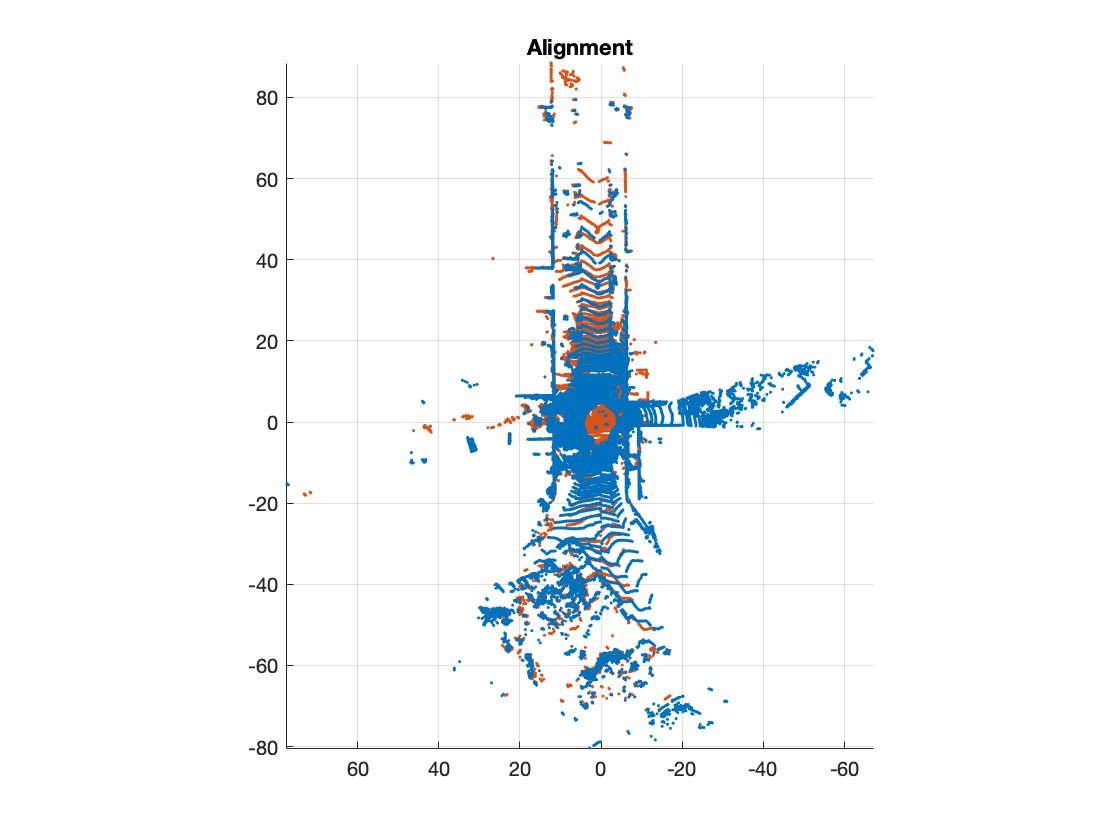}}	
	\subfloat[Scene 2. ISS+our desc]{
		\includegraphics[width=.25\linewidth]{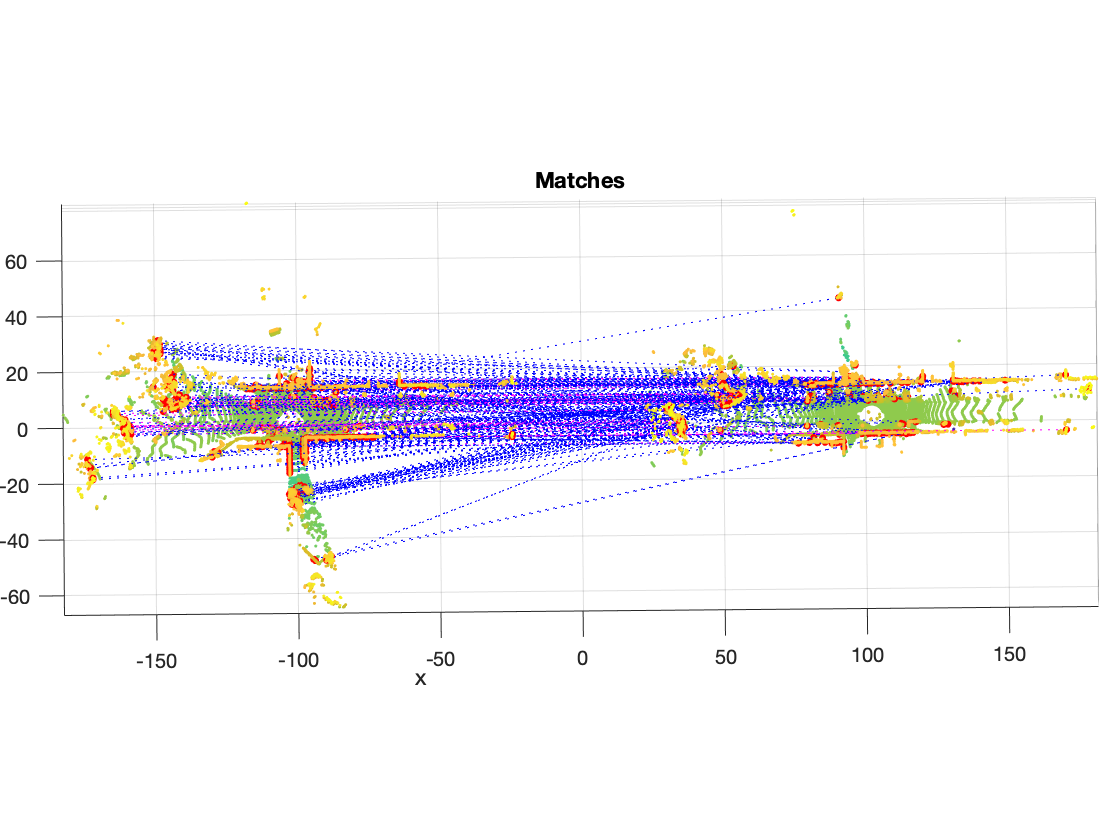}}
	\subfloat[Scene 2. 3DFeatNet kpt+our desc]{
		\includegraphics[width=.25\linewidth]{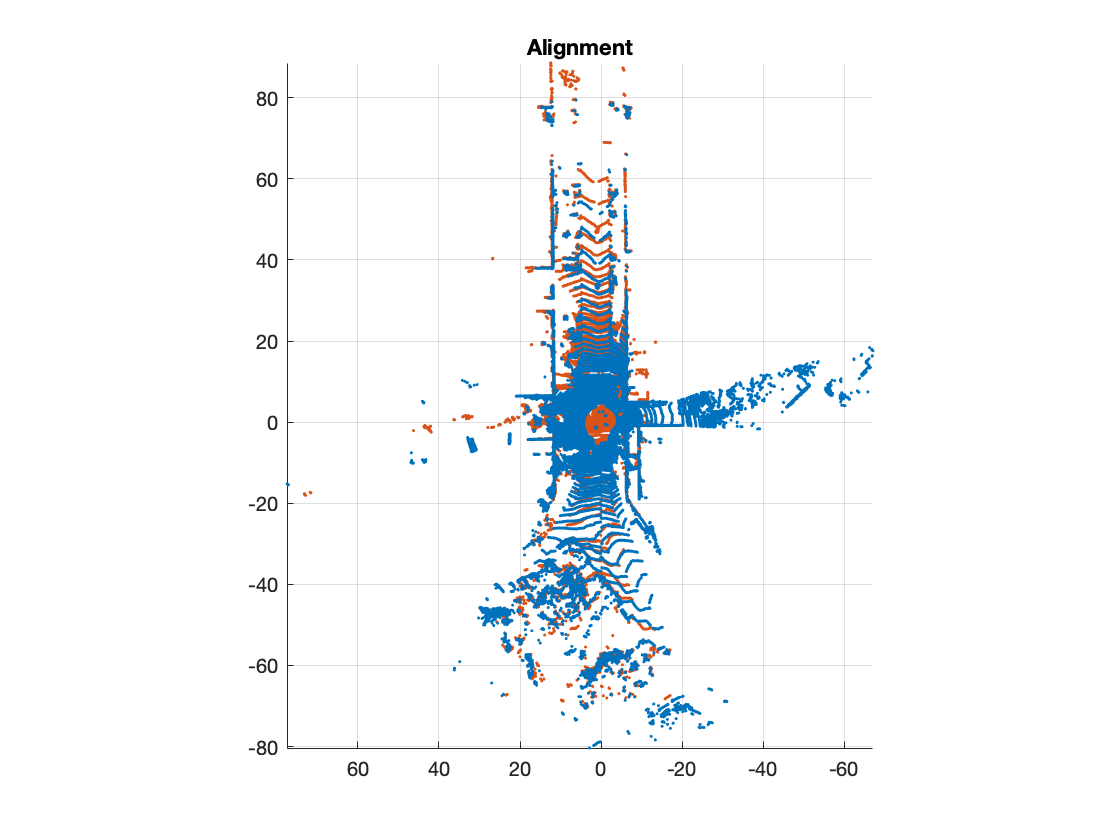}}
	\caption{KITTI data geometric registration. Top view. Blue dot line shows the matched pairs.}
	\label{fig:match_kitti}
\end{figure*}

\subsubsection{Geometric Registration}

With ISS keypoint and 3DFeatNet keypoint, we evaluate the descriptor algorithm on geometric registration. The registration is with the nearest neighbor match and applying RANSAC for transformation estimation. The RANSAC iteration is limited to $10,000$ and adjusted with $99\%$ confidence. Then Relative Rotation Error (RRE), Related Translation Error (RTE) are computed with ground truth to evaluate the accuracy of registration. A success is decided when RTE$<$\unit[2]{m}, RRE$<$\unit[5]{deg.}. The speed of converging is reflected by average number of iterations.
Since we use the same datasets (Oxford and KITTI) as~\cite{yew20183dfeat}, we compare to the results from their table.

The evaluation on Oxford data is demonstrated in Table~\ref{tab::oxford_registration}. The first eight rows are fetched from~\cite{yew20183dfeat} and the last five rows are from our own experiments.

We observe that, firstly, except for PN++, the handcrafted descriptors cannot exceed the learned descriptors. Secondly, our unsupervised learned descriptor achieves the best result on RRE and the success rate with ISS, while it has similar RTE and average iterators as 3DFeatNet on 3DFeatNet keypoint. Thirdly, the 3DFeatNet keypoint provide better accuracy for our descriptor. Regarding the success rate and convergence speed, when 3DFeatNet keypoint is used, they (3DFeatNet Desc and our Desc) are close to our approach. However, when ISS keypoint is utilized, our descriptor works better.

An example of a registration is shown in Fig.~\ref{fig:match_oxford}. We observe that ISS keypoints are distributed over whole point clouds while 3DFeatNet keypoints are mainly on the wall. For both keypoints, our descriptor matches accurately.

Then the algorithms are tested on other outdoor data, the KITTI dataset. The registration results are shown in Table~\ref{tab::kitti_registration}. The first six rows of  the results are fetched from~\cite{yew20183dfeat} and the last four rows are from our experiments. 

From the table, we observe that, firstly, ISS+our descriptor achieves best accuracy. And secondly, with 3DFeatNet keypoint, our model achieves as good results as the 3DFeatNet descriptor. A further example of a registration is in shown in  Fig.~\ref{fig:match_kitti}. For both ISS and 3DFeatNet keypoints, our descriptor performs well.

%

\subsection{Exploration}

As in~\cite{yew20183dfeat}, we also perform a descriptor matching test that doesn't use a keypoint detector.

\subsubsection{Random Point Descriptor Matching}
The data of descriptor matching is extracted from Oxford Robot Car Dataset at randomly selected locations. It consists of $30,000$ pairs of clusters with $4.0m$ radius. Half of them are matched pairs that are with matched frames and the other half are from point clouds at least $20m$ away. \cite{yew20183dfeat} uses a false-positive rate of $95\%$ recall.  

Please note that the clusters extracted are at a random location with the purpose of isolating the effect of keypoint detector. The keypoint locates in center of each cluster.

The result for our model is $66.33\%$, while the errors for SI, FPFH, USC, 3DMatch, PN++ and 3DFeatNet from~\cite{yew20183dfeat} are $68.51\%$, $54.13\%$, $91.59\%$, $59.77\%$, $38.49\%$, $50.57\%$ and $36.84\%$, respectively. Other than having equivalent performance as 3DFeatNet in the previous joint performance and geometric registration, our model only has $66.33\%$ recall.

So we consider that our learned model is not distinctive on random points as on keypoints. Thus we make further experiments on the given descriptor matching data to find if it performs on non-random keypoints. 

We use ISS to filter out the non-interested clusters pairs. For those $60,000$ clusters, we use the ISS keypoint detector with the same parameters in previous evaluation to detect keypoints in each cluster. Then we set a radius $r_{th}$ to threshold if there exists a keypoint that is in the sphere, which has radius $r_{th}$ on the center of cluster. If both clusters in a pair have keypoints in the sphere, this pair is kept for test.

The results are demonstrated in Table~\ref{tab::descriptorMatching}.

\begin{table}
	\centering
	\caption{Descriptor matching Error (\%). Lower is better.}
     \resizebox{\columnwidth}{!}{%
	  \begin{tabular}{|l||l|l|l|l|}
		\hline
		$r_{tj} (m)$& 0.1 & 0.2 &0.3 &0.4  \\ \hline 
		pairs(posi,neg) & 58(34,24)&178(103,75)&722(407,315)&1601(850,751) \\ \hline
		\hline
		3DFeatNet&  37.5   & 28    &\textbf{31.43} &\textbf{35.955}\\ \hline
		 Our&\textbf{29.17}&\textbf{13.33}&53.65&56.99
		 \\ \hline
	  \end{tabular}
	  \label{tab::descriptorMatching}
       }
\end{table}
We find that a large amount of clusters are not good keypoints. With small $r_{th}=0.1m$, $0.2m$, our model works better. However, when keypoints are farther to center, with $r_{th}$ larger, our model again does not works well.

Therefore, the use of our learned descriptor should be embedded with a key-point detector, but not with random points.

\section{Conclusion and Future Work}

In this paper, we proposed a self-supervised learning model to learn a point set local descriptor. Borrowing the high efficiency of the one-step registration model CF, a transformation is predicted with a learned descriptor function.
Thus, with a CF module as the last layer, only one point cloud is needed to be fed in the network for training. Using the same network body as 3DFeatNet, our model is much easier to train, because this self-supervised method does not require any manual labor on annotation and, without any pre-training, can converge with a higher learning rate, requiring far fewer iterations. Our experimental evaluation showed that our descriptor achieves equivalent performance on precision and geometric registration as the 3DFeatNet Descriptor. 

Needless to say, a lot of work remains to be done. As future work we want to more deeply explore the parameter space of the descriptor model to improve the descriptors of random points. Furthermore, we want to train and test our method with indoor LIDAR and RGBD datasets, that could then also use RGB information in the descriptor.


\IEEEtriggeratref{16}
\bibliographystyle{IEEEtran}
\bibliography{reference}

\end{document}